\pdfoutput=1

\documentclass[11pt]{article}

\usepackage[final]{acl}
\usepackage{booktabs} 
\usepackage{times}
\usepackage{latexsym}
\usepackage{multirow}
\usepackage{amsmath}
\usepackage{amsfonts}
\usepackage{arydshln}
\usepackage{listings}
\usepackage[T1]{fontenc}
\usepackage{svg}

\usepackage[utf8]{inputenc}

\usepackage{microtype}

\usepackage{inconsolata}

\usepackage{graphicx}
\usepackage{subcaption}
\usepackage{enumitem}
\title{Predicting Rewards Alongside Tokens: Non-disruptive Parameter Insertion for Efficient Inference Intervention in Large Language Model}

\author{Chenhan Yuan\textsuperscript{12}\thanks{Work done during internship at Alibaba Group}, Fei Huang\textsuperscript{2}\thanks{Corresponding author}, Ru Peng\textsuperscript{2}\\
\textbf{Keming Lu\textsuperscript{2}, Bowen Yu\textsuperscript{2}, Chang Zhou\textsuperscript{2}, Jingren Zhou\textsuperscript{2}}\\ 
\textsuperscript{1}The University of Manchester, Manchester, UK \hspace{0.25cm}  \textsuperscript{2}Alibaba Group, Hangzhou, China\\
chenhan.yuan@manchester.ac.uk \\
\{feihu.hf,rupeng.rp,lukeming.lkm,yubowen.ybw,ericzhou.zc,jingren.zhou\}@alibaba-inc.com
}

\begin{document}
\maketitle
\begin{abstract}
Transformer-based large language models (LLMs) exhibit limitations such as generating unsafe responses, unreliable reasoning, etc. 
Existing inference intervention approaches attempt to mitigate these issues by finetuning additional models to produce calibration signals (such as rewards) that guide the LLM's decoding process. 
However, this solution introduces substantial time and space overhead due to the separate models required. 
This work proposes N\textbf{O}n-disrup\textbf{t}ive parame\textbf{t}ers ins\textbf{er}tion (\textbf{Otter}), inserting extra parameters into the transformer architecture to predict calibration signals along with the original LLM output. Otter offers state-of-the-art performance on multiple demanding tasks while saving up to 86.5\% extra space and 98.5\% extra time. Furthermore, Otter seamlessly integrates with existing inference engines, requiring only a one-line code change, and the original model response remains accessible after the parameter insertion. Our code is publicly available at \url{https://github.com/chenhan97/Otter}
\end{abstract}

\section{Introduction}
Transformer-based Large Language Models (LLM) have demonstrated remarkable prowess in a wide range of natural language processing tasks with human-like proficiency, including text generation~\cite{qwen, qwen2}, translation~\cite{wu2023brief}, summarization~\cite{vassiliou2023summarygpt, yuan2023zero,zhang2023prompting,vaswani2017attention}, etc. However, extensive inspections from multiple directions have revealed LLMs are not flawless. For instance, LLM occasionally produces unsafe or toxic outputs~\cite{deng2023multilingual,khanov2024args}, and in reasoning capabilities, yielding unreliable results for mathematical proofs~\cite{yu2024metamath}.
\begin{figure}[t]
    \centering
\includegraphics[width=\columnwidth]{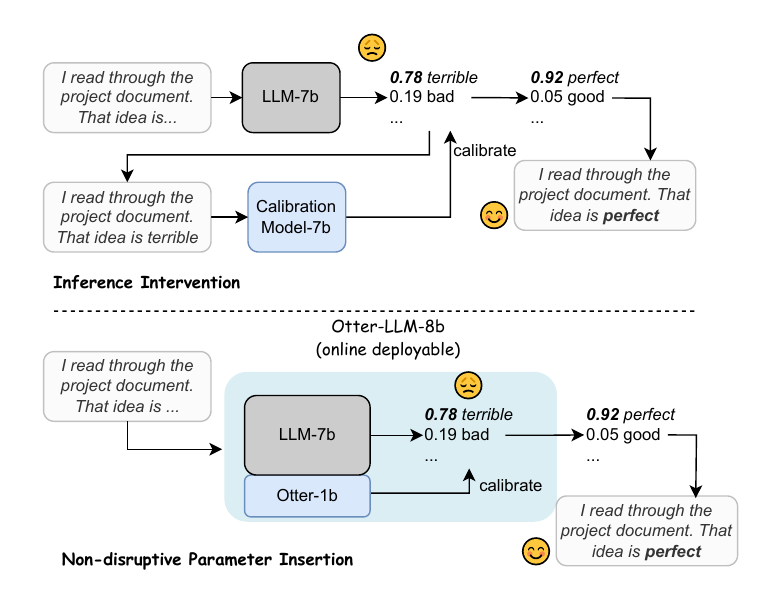}
    \caption{Comparison of inference intervention methods with and without Otter for harmless response generation. By inserting parameters into the frozen LLM, Otter significantly reduces space and time costs, while enabling seamless online deployment.}
    \label{fig:intro}
\end{figure}

To address the above limitations of LLMs, a series of approaches have been proposed, which can be broadly categorized into two directions -- \emph{finetuning} and \emph{inference intervention}. Finetuning methods leverage techniques such as knowledge graphs~\cite{10.1145/3589334.3645376} and data augmentation~\cite{yu2024metamath,bianchi2024safetytuned} to curate vast amounts of high-quality data to align LLMs with preferred responses. Nevertheless, prior research has highlighted that continual pretraining or finetuning can lead to unintended consequences, such as catastrophic forgetting~\cite{jian2022embedding,pmlr-v234-zhai24a,lu2023onlinedpo}, and hallucination~\cite{lin2023speciality}. These side effects can severely undermine the overall performance and reliability of LLMs. To avoid associated side effects, plug-and-play inference intervention as an alternative has been proposed. 
The method employs a calibration model to produce signals (such as rewards), which are used to calibrate the decoding tokens during LLM inference. When the intervention is not required, the calibration model can be unplugged, allowing the LLM to return to its original output. Benefiting from this, inference intervention achieves reliable reasoning, such as process-supervised reward models in mathematical reasoning~\cite{liu2021dexperts, yu2023outcome}, reduces bias and harmful responses with reward-guided search~\cite{khanov2024args}. Therefore, inference intervention paves a promising way to enhance LLMs' capabilities. 

Regretfully, the success of inference intervention heavily relies on training additional calibration models, which are typically large-scale models, and causes the space complexity to increase drastically~\cite{liu2021dexperts, khanov2024args}. In addition, as shown in Fig.~\ref{fig:intro}, the inference time is generally longer because the original model's output is re-used as input for the calibration model, increasing the times of transformer forward passes~\cite{yu2023outcome}. This introduces substantial space and time overhead, making it challenging to deploy these methods online. Motivated by the powerful knowledge inherent in LLMs, training an entirely new model is not necessary. Instead, it is reasonable to utilize this knowledge by re-using the original parameters. Specifically, our objective is to introduce a small set of additional parameters into the original LLM, allowing it to simultaneously generate accompanying output (either reward prediction or language modeling) alongside its primary output. This accompanying output should behave the same as the additional models in inference intervention methods.

Introducing additional parameters into LLMs has been indeed widely used in parameter-efficient finetuning/adapting methods (PEFT), such as QLoRA and LoRA~\cite{hu2022lora, NEURIPS2023_1feb8787}. These methods work by adding adapted parameters to the original frozen parameters, i.e., $h=(W+\Delta W)x$, where only the added $\Delta W$ is trainable. However, by directly adding parameters to the original parameters, LoRA alters all model outputs. Consequently, LoRA is not suitable for inference intervention methods, where the original model outputs (i.e., logits) and a calibration signal are required simultaneously.

In this work, we propose a N\textbf{O}n-disrup\textbf{t}ive parame\textbf{t}ers ins\textbf{er}tion (\textbf{Otter}) method for inference intervention in LLMs. The key idea behind Otter is to concatenate the added trainable parameters across all components of the transformer, including the multi-head attention layer and the feed-forward neural network layer. By passing through all layers, the final extended hidden state can be mapped and utilized as an inference intervention signal. Compared with existing work, this study offers several contributions:
\begin{enumerate}[leftmargin=1.5em]
    \item \textbf{SOTA performance with lowest overhead}. Otter saves up to \textbf{86.5\%} extra space and \textbf{98.5\%} extra time while obtaining comparable performance to \textbf{state-of-the-art} inference intervention methods on \textbf{three high-demanding tasks}: \emph{generation detoxification}, \emph{preference alignment}, and \emph{inference acceleration}.
    \item \textbf{Seamless integration}.
    Otter can integrate new parameters into the existing model so that the whole model can utilize the existing inference engine for efficient decoding with only \textbf{one line} of code modification.
    \item \textbf{Retain raw model response}. 
    The original language model's output is always accessible alongside the intervention signal. When the intervention is not required, Otter will not interfere with the original language model's output, preventing unexpected performance degradation. 
    
\end{enumerate}

\begin{figure*}[t]
\includegraphics[width=0.95\textwidth]{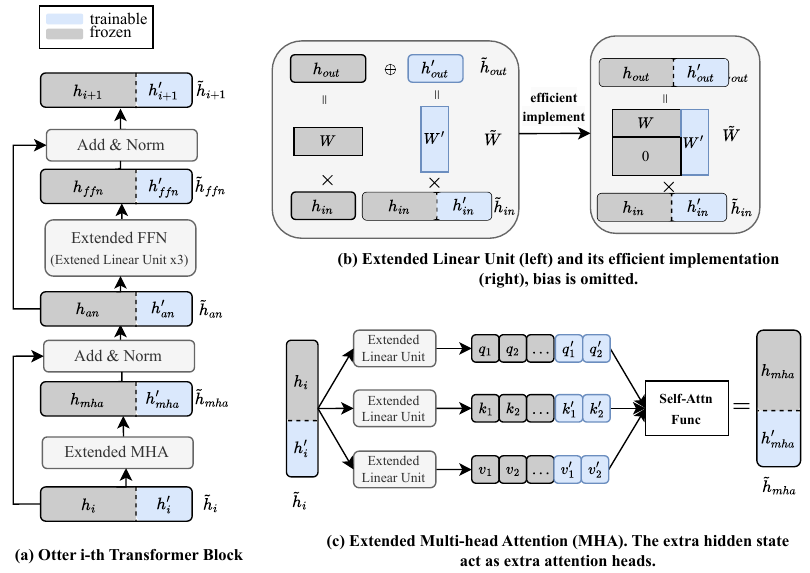}
\caption{The Otter architecture. Grey denotes frozen parameters while blue is trainable.}
\label{fig:otter}
\end{figure*}

\section{Related Works}
\subsection{Inference Intervention for LLM}
Existing inference intervention methods employ fine-tuned auxiliary models to guide LLM generation. The OVM~\cite{yu2023outcome} model selects top-k generations per step, improving mathematical reasoning. ARGS~\cite{khanov2024args} fine-tunes a reward model to calibrate token probabilities to generate more harmless responses. DEXP~\cite{liu2021dexperts} uses side models mimicking toxic-discriminative chatbots to guide non-toxic generation. \citet{leviathan2023fast} fine-tunes smaller LLMs to generate distributions resembling larger LLMs, enabling parallel candidate token generation for faster inference.
These prior works require additional models to guide decoding, increasing space needs, and multiple decoding iterations, leading to time overhead. Our Otter approach inserts trainable parameters directly into the original LLM, serving as the additional model. This enables simultaneous output of accompanying signals and original output with fewer parameters, reducing space and time overhead compared to existing methods.

\subsection{Parameter-Efficient Finetuning for LLM}
As pre-trained language models (PLMs) continue to grow in size, fine-tuning them becomes computationally infeasible. To address this issue, previous studies only fine-tuned a few parameters of PLMs, enabling comparable performance to full fine-tuning. These parameter-efficient fine-tuning methods can be categorized as follows:
\textbf{i)-Low-rank adaption} introduces a small number of new parameters that modulate the original PLM parameters. LoRA ~\cite{hu2022lora} proposes adding a low-rank adaptor to fine-tune LLMs. Subsequent QLoRA~\cite{NEURIPS2023_1feb8787} and AdaLoRA~\cite{zhang2023adaptive}, aim to further improve performance by quantizing the PLM and adaptively allocating parameter budgets, respectively. \textbf{ii)-Prompt-tuning methods} freeze the PLM and design task-specific prompts, with only the prompt-related parameters being fine-tuned, such as soft prompts~\cite{lester-etal-2021-power} and prefix-tuning~\cite{li-liang-2021-prefix}. \textbf{iii)-Adapters} also efficiently transfer knowledge for downstream tasks, with AdapterFusion~\cite{pfeiffer2021adapterfusion} combining knowledge from multiple source tasks, K-Adaptor~\cite{wang2021k} infusing different knowledge types, and AdapterHub~\cite{pfeiffer2020adapterhub} integrate pre-trained adapters across tasks and languages.
These methods aim to modify the model's output to improve task performance, which may also cause unexpected issues like hallucinations~\cite{lin2023speciality}. However, Otter overcomes these problems by expanding the weight matrices. This makes the new model a larger version of the original, adding capabilities without affecting the original output.

\section{Method}
We introduce Otter for predicting calibration signals alongside tokens in transformer-based LLMs without disrupting their original output and knowledge. Our method involves inserting new trainable parameters into the feed-forward network (FFN) and multi-head attention (MHA) layers of the transformer architecture. Specifically, for $i$-th block of the transformer, we expand the original hidden state $h_i$ into $\tilde{h}_i = [h_i, h'_i]$, thereby making it possible to predict the inference intervention signals based on the last layer $h'_n$. 

\subsection{FFN Layer Adaption}
The FFN layer in transformer models typically consists of three linear mappings, as defined in Equation~\ref{eq:ffn}. 
\begin{equation}
\begin{aligned}
    h_g &= W_gh_{an}+b_g \in \mathbb{R}^{d_{inner}} \\
    h_u &= W_uh_{an} + b_u \in \mathbb{R}^{d_{inner}}\\
    h_{ffn} &= W_d(\sigma(h_g)*h_u) +b_d \in \mathbb{R}^{d_{inp}} 
\end{aligned}
\label{eq:ffn}
\end{equation} where $\sigma(\cdot)$ is an activation function and $h_{an} \in \mathbb{R}^{d_{inp}}$ is the hidden state. We define $d_{inp}$ and $d_{inner}$ as the input dim. and inner dim. of the FFN layer, respectively, and refer to them henceforth.

In order to insert new trainable parameters while keeping the original hidden states intact, we design a method to expand the weight and bias matrix for each linear projection in the FFN layer, as illustrated in Figure \ref{fig:otter}(b). Specifically, taking the first linear projection in FFN as an example, our method preserves the linear projection from $h_{an}$ to $h_{g}$ unchanged and creates a new linear projection from $\tilde{h}_{an}=[h_{an}, h'_{an}]$ to $h'_{g}$.
To implement it efficiently, we expand each original weight matrix $W_g$ by concatenating a frozen all-zero matrix $O$ and a trainable adaptation weight matrix $W'_g$. 
The incorporation of the all-zero matrix $O$ ensures that the initial output $h_g$ remains unaltered. Moreover, the entire computational process can be implemented in a manner identical to the original linear projection without requiring any modifications to the existing code. The only difference is the increased size of the input and output dimensions.


\subsection{Multi-head Attention Layer Adaption}
In the multi-head attention (MHA) layer of the transformer, for an input hidden state $h_i$, the output $h_j$ is computed as follows:
\begin{equation}
\begin{aligned}
    Q = W_Qh_i \quad K = W_K&h_i\quad V = W_Vh_i \\
    head_m = Attn(&q_m, k_m, v_m) \\
    h_{mha} = Concat(head_1&, \cdots, head_n)W_O
\end{aligned} \notag
\end{equation}
where the query $Q$, key $K$, and value $V$ are obtained by linearly projecting $h_i$, then split into multiple heads ($q_m$, $k_m$, $v_m$) for attention computation.

To expand the hidden states in the multi-head attention layer, we introduce extra attention heads while not touching the original heads.
As depicted in Figure~\ref{fig:otter}(c), we insert the new parameters into the mapping matrices $W_Q, W_K$, $W_V$, and $W_O$ using the same method described in the FFN layer adaptation section, where the outputs of projections are then split into original frozen attention heads, such as $q_1$, $k_1$, $v_1$, and added trainable attention heads, such as $q_1'$,$k_1'$,$v_1'$.
Then we apply the multi-head attention operation on them and subsequently concatenate the outputs from both the original and newly introduced attention heads at the end of this layer.
Note that, this adaption only extends the size of hidden states and the number of attention heads, where no code modification is required, therefore seamlessly integrating efficient attention implementation such as FlashAttention~\cite{dao2022flashattention} and PagedAttention~\cite{kwon2023efficient}.

\subsection{Layer Norm Regularization}
\label{sec:reg}
In Otter, although the original input and output remain unchanged within each block of the transformer model, the layer normalization operation uses the mean and variance of the whole hidden states for normalization, which may disrupt the original hidden states. Therefore, we strictly enforce the use of only the original hidden state ($h_{mha}$ or $h_{ffn}$) to compute the variance for layer normalization. This ensures that the model's output remains consistent with the original model's behavior. Following common LLM choice, the normalization process after the FFN layer is defined using RMSNorm~\cite{zhang2019rmsnorm} in the following equation:
\begin{align}
\text{RMSNorm}(\tilde{h}_{ffn}) = \frac{\tilde{h}_{ffn}}{\sqrt{\text{mean}(h_{ffn}^2) + \epsilon}} \cdot \gamma \notag
\end{align} where $\gamma$ is learnable affine transform weights and $\epsilon$ is a small constant for stability. Please note that this is the only part of the method that requires modification to the inference codes, and this modification is relatively straightforward and will not affect the inference performance. We show that only one-line modification is needed in Appendix~\ref{appen_code_modify}.

To ensure the training stability of this restriction, a regularization term is introduced in the training objective $\mathcal{L}$ during the fine-tuning stage. In addition to the task-specific loss, we minimize the variance and mean difference between the original hidden state $h_{i}$ and the entire hidden state $\tilde{h}_i$.
\begin{gather}
    \mathcal{L}_{reg} = \sum_{i\in \mathcal{N}}(\sqrt{\text{mean}(h_i^2) + \epsilon} - \sqrt{\text{mean}(\tilde{h}_i^2) + \epsilon} ) ^2 \notag \\
    \mathcal{L} = \mathcal{L}_{task} + \lambda \mathcal{L}_{reg} \notag
\end{gather}
$\mathcal{N}$ is the number of blocks in the transformer. $\mathcal{L}_{task}$ is the task-related loss, such as mean squared error loss or cross-entropy loss used in finetuning the intervention model.
\begin{table*}[]
\small
\centering
\begin{tabular}{lccccccc}
\toprule
\multirow{2}{*}{Method}               & \multirow{2}{*}{Avg. Reward $\uparrow$} & \multirow{2}{*}{Diversity $\uparrow$} & \multirow{2}{*}{Coherence $\uparrow$} & \multirow{2}{*}{Win-Tie(\%) $\uparrow$} & \multicolumn{2}{c}{Overhead $\downarrow$}           & \multirow{2}{*}{\# Params.} \\ \cmidrule{6-7}
 &   &&     & & Time& Space              &    \\ \midrule
Greedy          & 3.981          & 0.567     & 0.426     &  50            & 1.0x      & 1.0x   &6.74B   \\
Greedy+ARGS     & \bf 5.026          & 0.611     & 0.456     &  \bf 64.33   & 2.07x            & 2.02x & 13.49B  \\
\ \ + Task Head Only  &  4.215         & 0.553     &   0.447  &   53.28  & 1.02x            & 1.03x & 6.75B    \\
\ \ \bf+ Otter  &  4.916         & \bf 0.745     &     \bf 0.503  &   62.75  & 1.03x            & 1.26x   & 8.51B    \\
\midrule
Top-k           & 3.757          & 0.679     & 0.463     & 50            & 1.0x    &  1.0x   & 6.74B     \\
Top-k+ARGS & \bf 4.787          & 0.700     & 0.463   &  \bf54.33  &     2.14x            & 2.02x  & 13.49B    \\
\ \ + Task Head Only  &  3.915        & 0.694     &   0.477   &   50.23  & 1.02x            &  1.03x & 6.75B     \\
\ \ \bf+ Otter     &  4.694         &  \bf 0.839     &  \bf 0.537     &53.59     & 1.04x            & 1.26x  & 8.51B         \\
\midrule
Top-p           & 3.799          & 0.636     & 0.511     & 50            & 1.0x    &  1.0x   & 6.74B     \\
Top-p+ARGS & \bf 4.803          & 0.665     & 0.519   & \bf 53.97  &     2.13x            & 2.02x  & 13.49B    \\
\ \ + Task Head Only     &  3.857         &  0.649     &  0.515     &50.42     & 1.02x            & 1.03x  & 6.75B         \\ 
\ \ \bf+ Otter     &  4.743         &  \bf 0.788     &  \bf 0.595     &53.61     & 1.04x            & 1.26x  & 8.51B         \\ \bottomrule
\end{tabular}
\caption{The experimental results of helpful and harmless alignment task. The Win-Tie rate compares the performance of ARGS and Otter against the baseline by GPT-4. Otter achieves comparable alignment level and text quality (average reward, diversity, coherence, and Win-Tie rate) to ARGS, while reducing extra space by $(1-\frac{1.26-1}{2.02-1})=\textbf{74.5\%}$ and extra time by $(1-\frac{1.03-1}{2.07-1})=\textbf{97.2\%}$.}
\label{tab:args_main}
\end{table*}

\section{Experiments}
\label{sec:exp_start}
To comprehensively assess Otter's capabilities, we consider three tasks:  human preference alignment~\cite{khanov2024args}, detoxification~\cite{liu2021dexperts}, and inference speed-up~\cite{cai2024medusa}. The evaluation involves various language models, including Llama~\cite{touvron2023llama}, Vicuna~\cite{zheng2024judging}, and GPT-2~\cite{radford2019language}. 

In addition to task-specific evaluation metrics, we analyze the computational overhead in terms of space and time complexity. We define the \textbf{space overhead} as the ratio of GPU memory consumed by the modified model to that of the original model during inference. The \textbf{time overhead} is defined as the ratio of the time consumed by the modified model to the time consumed by the base model. All inference interventions are performed on a single A100 GPU with the Huggingface transformers library~\cite{wolf2019huggingface}. 


In every task, we introduce an ablation of Otter, denoted as \textbf{Task Head Only}, which solely adds a trainable linear layer on the top of the frozen LLM. This linear layer is then fine-tuned to predict the intervention signals used in each individual task.

\subsection{Reward-guided Search for Helpful and Harmless Alignment}
The objective of this task is to enhance the capability of LLM to generate responses that are helpful and harmless, aligning with human preferences. The reward-guided search-based method involves fine-tuning a reward model, which is trained to assign higher values to text that is deemed helpful and harmless. During the decoding process, the reward model is utilized to adjust the original LLM's probabilistic predictions, thereby improving the alignment of generated responses with human preferences. Hyperparameters and detailed definitions can be found in Appendix~\ref{appen_hh}.
\subsubsection{Setup}
We employ the state-of-the-art ARGS model \cite{khanov2024args} as the reward model. Llama-7b is used as both the base model and the reward model. It is noteworthy that the base model remains unchanged, and only the reward model is fine-tuned on the preference dataset, which provides a signal for the inference intervention.
\vspace{0.2em}

\noindent\textbf{Dataset:} Consistent with the ARGS approach, we leverage the large-scale Helpful and Harmless (HH-RLHF) dataset~\cite{bai2022training} to validate the proposed Otter method against ARGS. This dataset comprises 112,000 training samples and 12,500 test samples and is publicly available.
\vspace{0.2em}

\noindent\textbf{Evaluation Metrics:} Following ARGS, we adopt the following metrics in addition to measuring overhead: \textbf{Average Reward}: This metric is the mean of the rewards computed by a \textbf{pre-trained and fixed} ARGS reward model across all generations from the HH-RLHF test prompts. It quantifies the model's ability to generate preferred and desirable outputs. \textbf{Diversity}: A score indicates the diversity of text. Higher is better. \textbf{Coherence}: A score shows the coherence level of text. Higher is better.
\subsubsection{Results}
\textbf{Otter Significantly Reduces ARGS Overhead.} As an inference intervention method, ARGS can be deployed under both common decoding settings: greedy decoding, top-k decoding, and top-p decoding. As shown in Table~\ref{tab:args_main}, ARGS significantly improves the base model's preference value and response diversity and coherence. However, this achievement comes at a considerable cost. ARGS model doubles both the time and space consumption compared to the base model. For example, compared with base model greedy decoding, ARGS-greedy incurs 2.07 times the time cost and requires 2.02 times space overhead during inference. In contrast, the Otter model reduces the time overhead from 2.07x to 1.03x, yielding $1-\frac{1.03-1}{2.07-1}=97.2\%$ saving in additional time compared to ARGS-greedy. Similarly, Otter reduces the space overhead from 2.02x to 1.26x, resulting in a 74.5\% reduction in additional space over ARGS.

\vspace{0.2em}
\noindent\textbf{Otter exhibits comparable alignment quality to ARGS.} As demonstrated in Table~\ref{tab:args_main}, Topk+ARGS and our method achieve average rewards of 4.787 and 4.694, respectively, both surpassing the base Top-k sampling model. Despite its lower computational complexity, Otter attains a performance comparable to ARGS' reward-guided search task performance.
To mitigate potential biases introduced by the pretrained reward model, following the ARGS, we additionally employ a GPT-4-based evaluation approach for comparing response quality by measuring the win-tie rate. The prompt can be found in Appendix~\ref{appen_hh_gpt4}. As shown in Table~\ref{tab:args_main}, compared to the base model, both ARGS and Otter maintain similar winning-tie rates. For example, for all prompts where ARGS wins against the base model under greedy decoding, Otter can win 97.5\% of those prompts.

\subsection{Controlled Bi-Experts Generation for Reducing Toxicity}
The task aims to mitigate the potential generation of toxic responses from LLMs. The controlled decoding-time experts generation method~(DEXP, \citealp{liu2021dexperts}) addresses this issue by employing two additional models: an expert model trained to generate non-toxic text, and an anti-expert model trained to generate toxic text. During the inference process of the original model, the next token probability distributions generated by these two auxiliary models are utilized to calibrate the original model's token generation probabilities, thereby reducing the likelihood of generating toxic content. More hyperparameters details can be found in Appendix~\ref{appen_toxi}.
\subsubsection{Setup}
Following DEXP, we use GPT-2-large~\cite{radford2019language} as the base model.

\vspace{0.2em}
\noindent\textbf{Datasets:}
We follow the setup in DEXP and use the human-annotated comments from the Jigsaw Unintended Bias in Toxicity Classification Kaggle challenge to train Otter. we use the 10K non-toxic prompts sampled by DEXP from the RealToxicityPrompts dataset~\cite{gehman-etal-2020-realtoxicityprompts}.  

\vspace{0.2em}
\noindent\textbf{Evaluation Metrics:} Following previous work~\cite{liu2021dexperts,gururangan-etal-2020-dont}, we apply Perspective API to measure the toxicity level of responses.\footnote{https://github.com/conversationai/perspectiveapi} We evaluate generation fluency using the mean perplexity of generated continuations based on a larger pre-trained GPT-2 XL model. Generation diversity is measured by the mean number of distinct n-grams, normalized by text length, among 25 generations for each prompt. We report Dist-1, Dist-2, and Dist-3 scores for distinct uni-, bi-, and trigrams, respectively.

\subsubsection{Results}
As shown in Table~\ref{tab:exp}, Otter achieves similar or better performance compared to the original DEXP model. We explore two settings: 1) using only the anti-expert for calibration, and 2) using both the anti-expert and expert for calibration. In the anti-expert-only setting, DEXP achieves an averaged maximum toxicity of 0.352, while Otter performs better with a score of 0.314. Notably, Otter maintains a similar generation quality to DEXP, as demonstrated by comparable perplexity and diversity scores.

Importantly, while the (anti-)expert models in DEXP have the same size as the original model, resulting in a 3.05x space overhead and a 2.97x inference time overhead, Otter significantly reduces these overheads to 1.13x and 1.03x, respectively. This makes Otter a more efficient and scalable solution for detoxifying text generations.
\begin{table*}[]
\small
\centering
\begin{tabular}{lccccccccc}
\toprule
\multirow{2}{*}{Decoding Method} & \multicolumn{2}{c}{Toxicity $\downarrow$} & Fluency $\downarrow$     & \multicolumn{3}{c}{Diversity $\uparrow$} & \multicolumn{2}{c}{Overhead $\downarrow$} & \multirow{2}{*}{\# Params.} \\ \cmidrule{2-9} 
                       & Avg. max.  & Toxicity prob.  & output ppl. & Dist-1   & Dist-2   & Dist-3  & Space         & Time         \\ \midrule
Top-p                  & 0.527           & 0.520      & 25.45       & 0.58     & 0.85     &  0.85    & 1.0x           & 1.0x  & 774M        \\ \midrule
DEXP(anti) & 0.352 & 0.191 & \bf 52.02 & \bf 0.58 &\bf 0.80 & \bf 0.73 & 2.02x & 1.984x & 1.55B \\
\ \ + Task Head Only     &     0.414    &  0.285    &  59.25 &   0.57       &    0.79      &    0.71      &  1.01x        &  1.015x   & 776M        \\ 
\ \ \bf + Otter      &     \bf 0.318    &  \bf 0.167    &  56.73 &   0.55       &    0.79      &    0.70      &  1.13x   &   1.023x  & 971M         \\
\midrule 
DEXP               & 0.314           & 0.128      & \bf 32.41       & \bf 0.58     & \bf 0.84     & \bf 0.84    & 3.05x           & 2.972x &2.32B        \\
\ \ + Task Head Only &  0.389  &  0.252  &   44.81          &      0.56  &    0.82      &   0.81      & 1.01x             &   1.026x  & 777M         \\
\ \ \bf+ Otter         &  \bf 0.295  &  \bf  0.119  &   38.56          &      0.56  &    0.80      &   0.80      &  1.27x             &   1.031x   & 1.24B        \\
\bottomrule
\end{tabular}
\caption{ Experimental results on detoxifying text generations from GPT-2. DEXP(anti) employs only an anti-expert signal for response calibration. Otter saves $(1-\frac{1.27-1}{3.05-1})=\textbf{86.5\%}$ extra space and $(1-\frac{1.03-1}{2.97-1})=\textbf{98.5\%}$ extra time with comparable detoxification performance to DEXP (i.e., toxicity probability, fluency, and diversity).}
\label{tab:exp}
\end{table*}

\subsection{Speculative Decoding}
Speculative decoding is a method to speed up the inference process, which approximates specific subtasks with efficient small draft models and subsequently employs the original model to validate these approximations in parallel, thereby reducing the overall decoding time without compromising the model's output distribution~\cite{leviathan2023fast}. The hyperparameters and detailed speculative decoding task definition can be found in Appendix~\ref{appen_sd}.
\subsubsection{Setup}
Our target model is set to be Vicuna-7b. As a baseline, we implemented the original speculative decoding~\cite{leviathan2023fast} by finetuning TinyLlama~\cite{zhang2024tinyllama} as the draft model, denoted as \emph{Vicuna-draft}.

We also compare with \emph{Medusa} model, which inserts decoding heads into the original model instead of separate draft models for inference acceleration~\cite{cai2024medusa}. Specifically, these additional heads in Medusa and Otter predict the next tokens in parallel, making the process more efficient than the traditional speculative decoding method that predicts draft tokens sequentially. Note that Medusa is the \textbf{Task Head Only} baseline in this experiment.

\vspace{0.2em}
\noindent\textbf{Dataset:} For training the Otter parameters, we utilize the ShareGPT dataset~\cite{zheng2024judging}, a public website where users share conversations with ChatGPT. The dataset contains 60k training samples, which are mostly overlapped with training samples of Vicuna-7b model. For evaluation, following Medusa, we use MT-Bench~\cite{zheng2024judging}, a multi-turn, and comprehensive conversational-format benchmark.

\vspace{0.2em}
\noindent\textbf{Evaluation Metrics:} Following Medusa, we use three metrics: a) Average accepted length: the average number of tokens decoded per decoding step (1.0 for standard auto-regressive models). b) Time overhead: The ratio of time cost per modified model forward pass to the time cost by the base model. c) Speedup: the wall-time acceleration rate, where Speedup = Average accepted length / Time overhead.
\subsubsection{Results}
\textbf{Otter significantly outperforms Vicuna-draft and Medusa in terms of inference speed-up.} As shown in Table~\ref{tab:medusa_main}, compared to conventional speculative decoding (e.g., Vicuna-draft), Otter and Medusa significantly reduced the space and time overhead by reducing inference times and added parameters. Otter can achieve $\frac{2.72 - 1.0}{1.0}=172\%$ speed gain compared to the base model, yielding a 45.5\%/14.8\% gain compared to Vicuna-draft and Medusa, respectively. Although Otter introduced more parameters inside the transformer model, leading to slightly higher time and space overhead compared to Medusa, these additional parameters were well-integrated with the original parameters, resulting in a much higher average acceptance length and speedup. 


\begin{table}[]
\small
\centering
\begin{tabular}{lcccc}
\toprule
\multirow{2}{*}{Method} & \multicolumn{2}{c}{Overhead$\downarrow$} & \multirow{2}{*}{Acpt. Len.$\uparrow$} & \multirow{2}{*}{Speedup$\uparrow$} \\ \cmidrule{2-3}
                        & Space               & Time               &                                       &                                    \\ \midrule
Vicuna-base             & 1.0x                 & 1.0x                & 1.0                                   & 1.0x                                \\
Vicuna-draft            & \bf1.09x                & 1.54x               & 2.87                                 & 1.87x                              \\
Medusa                  & 1.21x                & \bf1.06x               & 2.52                                 & 2.37x                              \\
\bf Otter                   & 1.24x                & 1.07x               & \bf2.91                                 &\bf2.72x                              \\ \bottomrule
\end{tabular}
\caption{Speedup of Otter and baselines for speculative decoding. Acpt.Len. denotes the average accepted length predicted by the added decoding heads.}
\label{tab:medusa_main}
\end{table}

\begin{table*}[]
\small
\centering
\begin{tabular}{cccccccc}
\toprule
\multicolumn{2}{c}{FFN} & \multirow{2}{*}{\# AttnHead} & \multicolumn{2}{c}{Overhead $\downarrow$} & \multirow{2}{*}{Accept length$\uparrow$} & \multirow{2}{*}{Speedup$\uparrow$} & \multirow{2}{*}{\# Params} \\ \cline{1-2} \cmidrule{4-5}
input dim  & inner dim  &                              & Time         & Space         &                                &                          &                            \\ \midrule 
128 & 256 & 4 & 1.07 & 1.24 & 2.91 & 2.72x &  7.37B\\ \midrule
128 & 0     &         8    &                  1.07    & 1.22     &   2.83    &  2.64x  &   7.60B                \\
256 & 0    &          6   &               1.07   & 1.23&      2.85         &  2.66x  & 7.67B                   \\
512 & 0    &    4    &     1.07    &1.25     &       \bf 2.88     &   \bf 2.69x    & 7.98B     \\
128 & 344    &    0    &    1.08  &\bf 1.14       &     2.60     &  2.41x & 7.02B         \\ \bottomrule
\end{tabular}
\caption{The comparisons of efficient Otter parameters insertion in FFN vs. MHA. Input and Inner dim. are inserted dims to the FFN layer. \# AttnHead is the number of inserted attention heads. For the same time overhead, inserting parameters into the MHA layer achieves a \textbf{19.9\%} speedup gain over FFN layer insertion in speculative decoding.}
\label{tab:medusa_comp}
\end{table*}
\begin{figure*}[h]
    \centering
    \includegraphics[width=\textwidth]{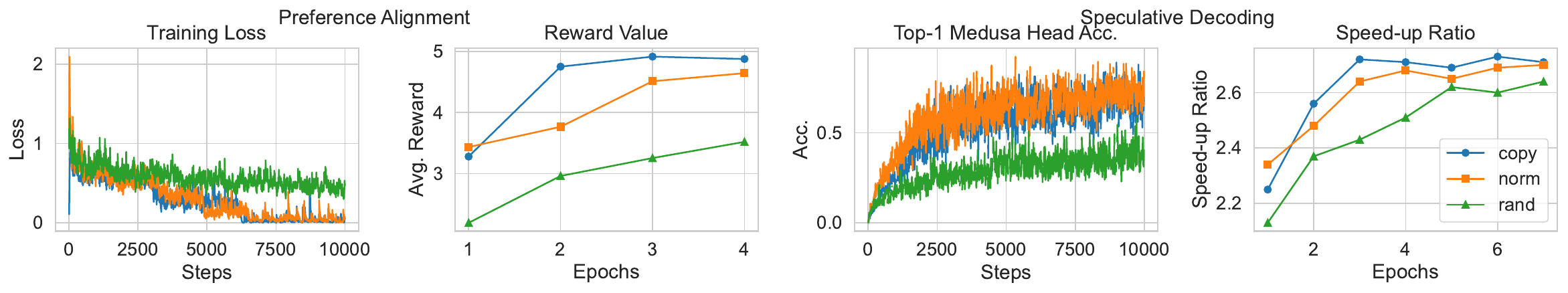}
    \caption{The comparisons of initialization methods' effectiveness on speculative decoding and preference alignment. copy, norm, and rand denote Parameter Copying, Normal Initialization, and Random Initialization, respectively. Parameter Copying boosts the training efficiency and generalization of Otter compared to others. The loss in the preference alignment task reflects the training of the reward model. Higher average reward values in this task indicate better alignment. In speculative decoding, the Top-1 medusa head Acc. measures the average next token prediction accuracy of the first medusa decoding head during training. Higher accuracy corresponds to a higher acceleration ratio. The speed-up ratio in this task quantifies the acceleration achieved against the base model, evaluated at each training epoch.}
    \label{fig:init}
\end{figure*}

\subsection{FFN vs. MHA: Efficient Otter Insertion}
Otter can be inserted into the FFN layer or the MHA layer of the transformer architecture. This raises an intriguing research question: \emph{which parameter group, FFN vs. MHA, is more efficient for the application of the Otter method?}  To investigate this issue, we carefully control the number of inserted parameters such that the time overhead remains constant in speculative decoding. As illustrated in Table~\ref{tab:medusa_comp}, our experiments revealed that for this task, adding parameters to the MHA layer significantly improved inference speed compared to augmenting the FFN layer. These findings suggest that for better alignment with the original model in speculative decoding, the multi-head attention component plays a more crucial role than the FFN layer. In contrast to previous research that posited that expanding the FFN parameters should be prioritized as the FFN layer is responsible for preserving knowledge~\cite{de2021editing}, this observation provides an alternative insight into parameter extension efficiency.
\subsection{Otter Initialization}
Through our experiments, we discovered the critical importance of investigating how different parameter initialization techniques affect Otter's overall performance.
We experimented with three initialization approaches: Random Initialization, Normal Initialization, and parameter Copying.
\begin{itemize}[leftmargin=1.5em]
    \item \textbf{Random Initialization:} parameters were initialized by sampling uniformly from the range (-0.5, 0.5).
    \item \textbf{Normal Initialization:} parameters were initialized by sampling from normal distributions with means and variances matching those of the original model parameter groups.
    \item \textbf{Parameter Copying:} parameters were initialized by randomly copying from the original model. For inserted FFN dimensions, parameters were copied from the original FFN layer and truncated as needed. For the MHA layer, if multiple new heads were inserted, each head is sampled from the original.
\end{itemize}

As illustrated in Fig.~\ref{fig:init}, both Normal Initialization and Parameter Copying methods outperformed Random Initialization on speculative decoding and preference alignment tasks. This outcome is expected, as the former two methods leverage the dense information from the original model parameters, reducing the likelihood of converging to local minima during training.
Notably, while Normal Initialization and Parameter Copying achieved similar performance on the training set, the Otter model initialized with Parameter Copying exhibited better generalization capabilities, as evidenced by its superior performance on the validation set. In summary, \emph{parameter copying is the most promising strategy among the three initialization methods}.

\section{Conclusion}
Otter concatenates a small set of trainable parameters to the transformer architecture, enabling additional output for inference intervention while preserving the original model's behavior and computational efficiency. Experiments demonstrate Otter's efficacy in aligning outputs with human preferences, mitigating toxicity in text generation, and improving inference speed—achieving these enhancements with significantly less overhead than existing methods. As an extension of the same model architecture, Otter can seamlessly integrate with existing infrastructure, providing an appealing solution for enhancing LLM performance efficiently for the community.
\section{Limitations}
While the Otter method offers a promising approach to non-disruptive parameter insertion for efficient inference intervention in large language models (LLMs), several aspects warrant further consideration. Firstly, despite the fact that Otter and other intervention methods can reduce harmful or unreliable generated responses, there is still a risk that the original LLM may generate such responses, and these methods cannot fully detect and calibrate them. This limitation may necessitate ongoing efforts to carefully pre-train the original LLM, such as cleaning the pre-training data and implementing safety alignment through reinforcement learning from human feedback (RLHF). Secondly, unlike other inference intervention methods that rely on separate calibration models, Otters insert parameters into the original model; thus, the hyperparameters such as the learning rate and initialization methods might differ from those of other approaches. This discrepancy may require extra effort in exploring the optimal hyperparameter settings specific to the Otters method.
\bibliography{custom}

\begin{thebibliography}{41}
\providecommand{\natexlab}[1]{#1}

\bibitem[{Bai et~al.(2023)Bai, Bai, Chu, Cui, Dang, Deng, Fan, Ge, Han, Huang, Hui, Ji, Li, Lin, Lin, Liu, Liu, Lu, Lu, Ma, Men, Ren, Ren, Tan, Tan, Tu, Wang, Wang, Wang, Wu, Xu, Xu, Yang, Yang, Yang, Yang, Yao, Yu, Yuan, Yuan, Zhang, Zhang, Zhang, Zhang, Zhou, Zhou, Zhou, and Zhu}]{qwen}
Jinze Bai, Shuai Bai, Yunfei Chu, Zeyu Cui, Kai Dang, Xiaodong Deng, Yang Fan, Wenbin Ge, Yu~Han, Fei Huang, Binyuan Hui, Luo Ji, Mei Li, Junyang Lin, Runji Lin, Dayiheng Liu, Gao Liu, Chengqiang Lu, Keming Lu, Jianxin Ma, Rui Men, Xingzhang Ren, Xuancheng Ren, Chuanqi Tan, Sinan Tan, Jianhong Tu, Peng Wang, Shijie Wang, Wei Wang, Shengguang Wu, Benfeng Xu, Jin Xu, An~Yang, Hao Yang, Jian Yang, Shusheng Yang, Yang Yao, Bowen Yu, Hongyi Yuan, Zheng Yuan, Jianwei Zhang, Xingxuan Zhang, Yichang Zhang, Zhenru Zhang, Chang Zhou, Jingren Zhou, Xiaohuan Zhou, and Tianhang Zhu. 2023.
\newblock Qwen technical report.
\newblock \emph{arXiv preprint arXiv:2309.16609}.

\bibitem[{Bai et~al.(2022)Bai, Jones, Ndousse, Askell, Chen, DasSarma, Drain, Fort, Ganguli, Henighan et~al.}]{bai2022training}
Yuntao Bai, Andy Jones, Kamal Ndousse, Amanda Askell, Anna Chen, Nova DasSarma, Dawn Drain, Stanislav Fort, Deep Ganguli, Tom Henighan, et~al. 2022.
\newblock Training a helpful and harmless assistant with reinforcement learning from human feedback.
\newblock \emph{arXiv preprint arXiv:2204.05862}.

\bibitem[{Bianchi et~al.(2024)Bianchi, Suzgun, Attanasio, Rottger, Jurafsky, Hashimoto, and Zou}]{bianchi2024safetytuned}
Federico Bianchi, Mirac Suzgun, Giuseppe Attanasio, Paul Rottger, Dan Jurafsky, Tatsunori Hashimoto, and James Zou. 2024.
\newblock \href {https://openreview.net/forum?id=gT5hALch9z} {Safety-tuned {LL}a{MA}s: Lessons from improving the safety of large language models that follow instructions}.
\newblock In \emph{The Twelfth International Conference on Learning Representations}.

\bibitem[{Cai et~al.(2024)Cai, Li, Geng, Peng, Lee, Chen, and Dao}]{cai2024medusa}
Tianle Cai, Yuhong Li, Zhengyang Geng, Hongwu Peng, Jason~D Lee, Deming Chen, and Tri Dao. 2024.
\newblock Medusa: Simple llm inference acceleration framework with multiple decoding heads.
\newblock \emph{arXiv preprint arXiv:2401.10774}.

\bibitem[{Dao et~al.(2022)Dao, Fu, Ermon, Rudra, and R{\'e}}]{dao2022flashattention}
Tri Dao, Dan Fu, Stefano Ermon, Atri Rudra, and Christopher R{\'e}. 2022.
\newblock Flashattention: Fast and memory-efficient exact attention with io-awareness.
\newblock \emph{Advances in Neural Information Processing Systems}, 35:16344--16359.

\bibitem[{De~Cao et~al.(2021)De~Cao, Aziz, and Titov}]{de2021editing}
Nicola De~Cao, Wilker Aziz, and Ivan Titov. 2021.
\newblock Editing factual knowledge in language models.
\newblock In \emph{Proceedings of the 2021 Conference on Empirical Methods in Natural Language Processing}, pages 6491--6506.

\bibitem[{Deng et~al.(2023)Deng, Zhang, Pan, and Bing}]{deng2023multilingual}
Yue Deng, Wenxuan Zhang, Sinno~Jialin Pan, and Lidong Bing. 2023.
\newblock Multilingual jailbreak challenges in large language models.
\newblock In \emph{The Twelfth International Conference on Learning Representations}.

\bibitem[{Dettmers et~al.(2023)Dettmers, Pagnoni, Holtzman, and Zettlemoyer}]{NEURIPS2023_1feb8787}
Tim Dettmers, Artidoro Pagnoni, Ari Holtzman, and Luke Zettlemoyer. 2023.
\newblock \href {https://proceedings.neurips.cc/paper_files/paper/2023/file/1feb87871436031bdc0f2beaa62a049b-Paper-Conference.pdf} {Qlora: Efficient finetuning of quantized llms}.
\newblock In \emph{Advances in Neural Information Processing Systems}, volume~36, pages 10088--10115. Curran Associates, Inc.

\bibitem[{Gehman et~al.(2020)Gehman, Gururangan, Sap, Choi, and Smith}]{gehman-etal-2020-realtoxicityprompts}
Samuel Gehman, Suchin Gururangan, Maarten Sap, Yejin Choi, and Noah~A. Smith. 2020.
\newblock \href {https://doi.org/10.18653/v1/2020.findings-emnlp.301} {{R}eal{T}oxicity{P}rompts: Evaluating neural toxic degeneration in language models}.
\newblock In \emph{Findings of the Association for Computational Linguistics: EMNLP 2020}, pages 3356--3369, Online. Association for Computational Linguistics.

\bibitem[{Gururangan et~al.(2020)Gururangan, Marasovi{\'c}, Swayamdipta, Lo, Beltagy, Downey, and Smith}]{gururangan-etal-2020-dont}
Suchin Gururangan, Ana Marasovi{\'c}, Swabha Swayamdipta, Kyle Lo, Iz~Beltagy, Doug Downey, and Noah~A. Smith. 2020.
\newblock \href {https://doi.org/10.18653/v1/2020.acl-main.740} {Don{'}t stop pretraining: Adapt language models to domains and tasks}.
\newblock In \emph{Proceedings of the 58th Annual Meeting of the Association for Computational Linguistics}, pages 8342--8360, Online. Association for Computational Linguistics.

\bibitem[{Hu et~al.(2022)Hu, yelong shen, Wallis, Allen-Zhu, Li, Wang, Wang, and Chen}]{hu2022lora}
Edward~J Hu, yelong shen, Phillip Wallis, Zeyuan Allen-Zhu, Yuanzhi Li, Shean Wang, Lu~Wang, and Weizhu Chen. 2022.
\newblock \href {https://openreview.net/forum?id=nZeVKeeFYf9} {Lo{RA}: Low-rank adaptation of large language models}.
\newblock In \emph{International Conference on Learning Representations}.

\bibitem[{Jian et~al.(2022)Jian, Gao, and Vosoughi}]{jian2022embedding}
Yiren Jian, Chongyang Gao, and Soroush Vosoughi. 2022.
\newblock Embedding hallucination for few-shot language fine-tuning.
\newblock In \emph{Proceedings of the 2022 Conference of the North American Chapter of the Association for Computational Linguistics: Human Language Technologies}, pages 5522--5530.

\bibitem[{Khanov et~al.(2024)Khanov, Burapacheep, and Li}]{khanov2024args}
Maxim Khanov, Jirayu Burapacheep, and Yixuan Li. 2024.
\newblock \href {https://openreview.net/forum?id=shgx0eqdw6} {{ARGS}: Alignment as reward-guided search}.
\newblock In \emph{The Twelfth International Conference on Learning Representations}.

\bibitem[{Kwon et~al.(2023)Kwon, Li, Zhuang, Sheng, Zheng, Yu, Gonzalez, Zhang, and Stoica}]{kwon2023efficient}
Woosuk Kwon, Zhuohan Li, Siyuan Zhuang, Ying Sheng, Lianmin Zheng, Cody~Hao Yu, Joseph Gonzalez, Hao Zhang, and Ion Stoica. 2023.
\newblock Efficient memory management for large language model serving with pagedattention.
\newblock In \emph{Proceedings of the 29th Symposium on Operating Systems Principles}, pages 611--626.

\bibitem[{Lester et~al.(2021)Lester, Al-Rfou, and Constant}]{lester-etal-2021-power}
Brian Lester, Rami Al-Rfou, and Noah Constant. 2021.
\newblock \href {https://doi.org/10.18653/v1/2021.emnlp-main.243} {The power of scale for parameter-efficient prompt tuning}.
\newblock In \emph{Proceedings of the 2021 Conference on Empirical Methods in Natural Language Processing}, pages 3045--3059, Online and Punta Cana, Dominican Republic. Association for Computational Linguistics.

\bibitem[{Leviathan et~al.(2023)Leviathan, Kalman, and Matias}]{leviathan2023fast}
Yaniv Leviathan, Matan Kalman, and Yossi Matias. 2023.
\newblock Fast inference from transformers via speculative decoding.
\newblock In \emph{International Conference on Machine Learning}, pages 19274--19286. PMLR.

\bibitem[{Li and Liang(2021)}]{li-liang-2021-prefix}
Xiang~Lisa Li and Percy Liang. 2021.
\newblock \href {https://doi.org/10.18653/v1/2021.acl-long.353} {Prefix-tuning: Optimizing continuous prompts for generation}.
\newblock In \emph{Proceedings of the 59th Annual Meeting of the Association for Computational Linguistics and the 11th International Joint Conference on Natural Language Processing (Volume 1: Long Papers)}, pages 4582--4597, Online. Association for Computational Linguistics.

\bibitem[{Lin et~al.(2023)Lin, Tan, Lin, Zheng, Pi, Zhang, Diao, Wang, Zhao, Yao et~al.}]{lin2023speciality}
Yong Lin, Lu~Tan, Hangyu Lin, Zeming Zheng, Renjie Pi, Jipeng Zhang, Shizhe Diao, Haoxiang Wang, Han Zhao, Yuan Yao, et~al. 2023.
\newblock Speciality vs generality: An empirical study on catastrophic forgetting in fine-tuning foundation models.
\newblock \emph{arXiv preprint arXiv:2309.06256}.

\bibitem[{Liu et~al.(2021)Liu, Sap, Lu, Swayamdipta, Bhagavatula, Smith, and Choi}]{liu2021dexperts}
Alisa Liu, Maarten Sap, Ximing Lu, Swabha Swayamdipta, Chandra Bhagavatula, Noah~A Smith, and Yejin Choi. 2021.
\newblock Dexperts: Decoding-time controlled text generation with experts and anti-experts.
\newblock In \emph{Proceedings of the 59th Annual Meeting of the Association for Computational Linguistics and the 11th International Joint Conference on Natural Language Processing (Volume 1: Long Papers)}, pages 6691--6706.

\bibitem[{Lu et~al.(2024)Lu, Yu, Huang, Yang, Lin, and Zhou}]{lu2023onlinedpo}
Keming Lu, Bowen Yu, Fei Huang, Fan Yang, Runji Lin, and Chang Zhou. 2024.
\newblock Online merging optimizers for boosting rewards and mitigating tax in alignment.
\newblock \emph{arXiv preprint arXiv:2405.17931}.

\bibitem[{Pfeiffer et~al.(2021)Pfeiffer, Kamath, R{\"u}ckl{\'e}, Cho, and Gurevych}]{pfeiffer2021adapterfusion}
Jonas Pfeiffer, Aishwarya Kamath, Andreas R{\"u}ckl{\'e}, Kyunghyun Cho, and Iryna Gurevych. 2021.
\newblock Adapterfusion: Non-destructive task composition for transfer learning.
\newblock In \emph{Proceedings of the 16th Conference of the European Chapter of the Association for Computational Linguistics: Main Volume}, pages 487--503.

\bibitem[{Pfeiffer et~al.(2020)Pfeiffer, R{\"u}ckl{\'e}, Poth, Kamath, Vuli{\'c}, Ruder, Cho, and Gurevych}]{pfeiffer2020adapterhub}
Jonas Pfeiffer, Andreas R{\"u}ckl{\'e}, Clifton Poth, Aishwarya Kamath, Ivan Vuli{\'c}, Sebastian Ruder, Kyunghyun Cho, and Iryna Gurevych. 2020.
\newblock Adapterhub: A framework for adapting transformers.
\newblock In \emph{Proceedings of the 2020 Conference on Empirical Methods in Natural Language Processing: System Demonstrations}, pages 46--54.

\bibitem[{Radford et~al.(2019)Radford, Wu, Child, Luan, Amodei, Sutskever et~al.}]{radford2019language}
Alec Radford, Jeffrey Wu, Rewon Child, David Luan, Dario Amodei, Ilya Sutskever, et~al. 2019.
\newblock Language models are unsupervised multitask learners.
\newblock \emph{OpenAI blog}, 1(8):9.

\bibitem[{Su et~al.(2022)Su, Lan, Wang, Yogatama, Kong, and Collier}]{su2022contrastive}
Yixuan Su, Tian Lan, Yan Wang, Dani Yogatama, Lingpeng Kong, and Nigel Collier. 2022.
\newblock A contrastive framework for neural text generation.
\newblock \emph{Advances in Neural Information Processing Systems}, 35:21548--21561.

\bibitem[{Touvron et~al.(2023)Touvron, Martin, Stone, Albert, Almahairi, Babaei, Bashlykov, Batra, Bhargava, Bhosale et~al.}]{touvron2023llama}
Hugo Touvron, Louis Martin, Kevin Stone, Peter Albert, Amjad Almahairi, Yasmine Babaei, Nikolay Bashlykov, Soumya Batra, Prajjwal Bhargava, Shruti Bhosale, et~al. 2023.
\newblock Llama 2: Open foundation and fine-tuned chat models.
\newblock \emph{arXiv preprint arXiv:2307.09288}.

\bibitem[{Vassiliou et~al.(2023)Vassiliou, Papadakis, and Kondylakis}]{vassiliou2023summarygpt}
Giannis Vassiliou, Nikolaos Papadakis, and Haridimos Kondylakis. 2023.
\newblock Summarygpt: Leveraging chatgpt for summarizing knowledge graphs.
\newblock In \emph{European Semantic Web Conference}, pages 164--168. Springer.

\bibitem[{Vaswani et~al.(2017)Vaswani, Shazeer, Parmar, Uszkoreit, Jones, Gomez, Kaiser, and Polosukhin}]{vaswani2017attention}
Ashish Vaswani, Noam Shazeer, Niki Parmar, Jakob Uszkoreit, Llion Jones, Aidan~N Gomez, {\L}ukasz Kaiser, and Illia Polosukhin. 2017.
\newblock Attention is all you need.
\newblock \emph{Advances in neural information processing systems}, 30.

\bibitem[{Wang et~al.(2021)Wang, Tang, Duan, Wei, Huang, Ji, Cao, Jiang, and Zhou}]{wang2021k}
Ruize Wang, Duyu Tang, Nan Duan, Zhongyu Wei, Xuan-Jing Huang, Jianshu Ji, Guihong Cao, Daxin Jiang, and Ming Zhou. 2021.
\newblock K-adapter: Infusing knowledge into pre-trained models with adapters.
\newblock In \emph{Findings of the Association for Computational Linguistics: ACL-IJCNLP 2021}, pages 1405--1418.

\bibitem[{Wolf et~al.(2019)Wolf, Debut, Sanh, Chaumond, Delangue, Moi, Cistac, Rault, Louf, Funtowicz et~al.}]{wolf2019huggingface}
Thomas Wolf, Lysandre Debut, Victor Sanh, Julien Chaumond, Clement Delangue, Anthony Moi, Pierric Cistac, Tim Rault, R{\'e}mi Louf, Morgan Funtowicz, et~al. 2019.
\newblock Huggingface's transformers: State-of-the-art natural language processing.
\newblock \emph{arXiv preprint arXiv:1910.03771}.

\bibitem[{Wu et~al.(2023)Wu, He, Liu, Sun, Liu, Han, and Tang}]{wu2023brief}
Tianyu Wu, Shizhu He, Jingping Liu, Siqi Sun, Kang Liu, Qing-Long Han, and Yang Tang. 2023.
\newblock A brief overview of chatgpt: The history, status quo and potential future development.
\newblock \emph{IEEE/CAA Journal of Automatica Sinica}, 10(5):1122--1136.

\bibitem[{Yang et~al.(2024)Yang, Yang, Hui, Zheng, Yu, Zhou, Li, Li, Liu, Huang, Dong, Wei, Lin, Tang, Wang, Yang, Tu, Zhang, Ma, Yang, Xu, Zhou, Bai, He, Lin, Dang, Lu, Chen, Yang, Li, Xue, Ni, Zhang, Wang, Peng, Men, Gao, Lin, Wang, Bai, Tan, Zhu, Li, Liu, Ge, Deng, Zhou, Ren, Zhang, Wei, Ren, Liu, Fan, Yao, Zhang, Wan, Chu, Liu, Cui, Zhang, Guo, and Fan}]{qwen2}
An~Yang, Baosong Yang, Binyuan Hui, Bo~Zheng, Bowen Yu, Chang Zhou, Chengpeng Li, Chengyuan Li, Dayiheng Liu, Fei Huang, Guanting Dong, Haoran Wei, Huan Lin, Jialong Tang, Jialin Wang, Jian Yang, Jianhong Tu, Jianwei Zhang, Jianxin Ma, Jianxin Yang, Jin Xu, Jingren Zhou, Jinze Bai, Jinzheng He, Junyang Lin, Kai Dang, Keming Lu, Keqin Chen, Kexin Yang, Mei Li, Mingfeng Xue, Na~Ni, Pei Zhang, Peng Wang, Ru~Peng, Rui Men, Ruize Gao, Runji Lin, Shijie Wang, Shuai Bai, Sinan Tan, Tianhang Zhu, Tianhao Li, Tianyu Liu, Wenbin Ge, Xiaodong Deng, Xiaohuan Zhou, Xingzhang Ren, Xinyu Zhang, Xipin Wei, Xuancheng Ren, Xuejing Liu, Yang Fan, Yang Yao, Yichang Zhang, Yu~Wan, Yunfei Chu, Yuqiong Liu, Zeyu Cui, Zhenru Zhang, Zhifang Guo, and Zhihao Fan. 2024.
\newblock \href {https://arxiv.org/abs/2407.10671} {Qwen2 technical report}.
\newblock \emph{Preprint}, arXiv:2407.10671.

\bibitem[{Yu et~al.(2023)Yu, Gao, and Wang}]{yu2023outcome}
Fei Yu, Anningzhe Gao, and Benyou Wang. 2023.
\newblock Outcome-supervised verifiers for planning in mathematical reasoning.
\newblock \emph{arXiv preprint arXiv:2311.09724}.

\bibitem[{Yu et~al.(2024)Yu, Jiang, Shi, YU, Liu, Zhang, Kwok, Li, Weller, and Liu}]{yu2024metamath}
Longhui Yu, Weisen Jiang, Han Shi, Jincheng YU, Zhengying Liu, Yu~Zhang, James Kwok, Zhenguo Li, Adrian Weller, and Weiyang Liu. 2024.
\newblock \href {https://openreview.net/forum?id=N8N0hgNDRt} {Metamath: Bootstrap your own mathematical questions for large language models}.
\newblock In \emph{The Twelfth International Conference on Learning Representations}.

\bibitem[{Yuan et~al.(2023)Yuan, Xie, and Ananiadou}]{yuan2023zero}
Chenhan Yuan, Qianqian Xie, and Sophia Ananiadou. 2023.
\newblock Zero-shot temporal relation extraction with chatgpt.
\newblock \emph{arXiv preprint arXiv:2304.05454}.

\bibitem[{Yuan et~al.(2024)Yuan, Xie, Huang, and Ananiadou}]{10.1145/3589334.3645376}
Chenhan Yuan, Qianqian Xie, Jimin Huang, and Sophia Ananiadou. 2024.
\newblock \href {https://doi.org/10.1145/3589334.3645376} {Back to the future: Towards explainable temporal reasoning with large language models}.
\newblock In \emph{Proceedings of the ACM on Web Conference 2024}, WWW '24, page 1963–1974, New York, NY, USA. Association for Computing Machinery.

\bibitem[{Zhai et~al.(2024)Zhai, Tong, Li, Cai, Qu, Lee, and Ma}]{pmlr-v234-zhai24a}
Yuexiang Zhai, Shengbang Tong, Xiao Li, Mu~Cai, Qing Qu, Yong~Jae Lee, and Yi~Ma. 2024.
\newblock \href {https://proceedings.mlr.press/v234/zhai24a.html} {Investigating the catastrophic forgetting in multimodal large language model fine-tuning}.
\newblock In \emph{Conference on Parsimony and Learning}, volume 234 of \emph{Proceedings of Machine Learning Research}, pages 202--227. PMLR.

\bibitem[{Zhang et~al.(2023{\natexlab{a}})Zhang, Haddow, and Birch}]{zhang2023prompting}
Biao Zhang, Barry Haddow, and Alexandra Birch. 2023{\natexlab{a}}.
\newblock Prompting large language model for machine translation: A case study.
\newblock In \emph{International Conference on Machine Learning}, pages 41092--41110. PMLR.

\bibitem[{Zhang and Sennrich(2019)}]{zhang2019rmsnorm}
Biao Zhang and Rico Sennrich. 2019.
\newblock \href {https://proceedings.neurips.cc/paper/2019/hash/1e8a19426224ca89e83cef47f1e7f53b-Abstract.html} {Root mean square layer normalization}.
\newblock In \emph{Advances in Neural Information Processing Systems 32: Annual Conference on Neural Information Processing Systems 2019, NeurIPS 2019, December 8-14, 2019, Vancouver, BC, Canada}, pages 12360--12371.

\bibitem[{Zhang et~al.(2024)Zhang, Zeng, Wang, and Lu}]{zhang2024tinyllama}
Peiyuan Zhang, Guangtao Zeng, Tianduo Wang, and Wei Lu. 2024.
\newblock \href {https://arxiv.org/abs/2401.02385} {Tinyllama: An open-source small language model}.
\newblock \emph{Preprint}, arXiv:2401.02385.

\bibitem[{Zhang et~al.(2023{\natexlab{b}})Zhang, Chen, Bukharin, He, Cheng, Chen, and Zhao}]{zhang2023adaptive}
Qingru Zhang, Minshuo Chen, Alexander Bukharin, Pengcheng He, Yu~Cheng, Weizhu Chen, and Tuo Zhao. 2023{\natexlab{b}}.
\newblock \href {https://openreview.net/forum?id=lq62uWRJjiY} {Adaptive budget allocation for parameter-efficient fine-tuning}.
\newblock In \emph{The Eleventh International Conference on Learning Representations}.

\bibitem[{Zheng et~al.(2024)Zheng, Chiang, Sheng, Zhuang, Wu, Zhuang, Lin, Li, Li, Xing et~al.}]{zheng2024judging}
Lianmin Zheng, Wei-Lin Chiang, Ying Sheng, Siyuan Zhuang, Zhanghao Wu, Yonghao Zhuang, Zi~Lin, Zhuohan Li, Dacheng Li, Eric Xing, et~al. 2024.
\newblock Judging llm-as-a-judge with mt-bench and chatbot arena.
\newblock \emph{Advances in Neural Information Processing Systems}, 36.

\end{thebibliography}

\clearpage

\appendix
\section{Experiment Details}
We perform all Otter training on 4 NVIDIA A100 GPUs. Deepspeed ZeRO stage 2 is enabled to perform parallel training. 
\subsection{Helpful and Harmless Alignment}
\label{appen_hh}
\subsubsection{Hyperparameter setting}The hyperparameters employed for the Otter model on the helpful and harmless alignment task are summarized in Table~\ref{tab:appen_hh_hyper}. The regularization weight represents the significance of the added training loss objective, as described in Section~\ref{sec:reg}.
\begin{table}[]
\centering
\begin{tabular}{lc}
\toprule
Parameters            & Value \\ \midrule
\# of training epochs & 5     \\
Learning rate         & 5e-6  \\
Warm-up               & 0.01  \\
LayerNorm regularization $\lambda$ & 5     \\
FFN Otter input dim   & 256   \\
FFN Otter inner dim   & 512   \\
\# Otter attn heads    & 16    \\ \bottomrule
\end{tabular}
\caption{Summary of hyperparameter settings of Otter on helpful and harmless alignment}
\label{tab:appen_hh_hyper}
\end{table}
\subsubsection{ARGS Decoding}
The ARGS decoding technique calibrates the next token score $v$ (or probability) when generating subsequent tokens, utilizing the following equation:
\begin{equation}
    s(v,x_{<t})=LM(v|x_{<t})+w\cdot r([x_{<t},v])
\end{equation}Here, $LM(v|x_{<t})$ represents the original model output, $r(\cdot)$ is the reward model, $w$ is the weight assigned to the reward scalar, and $[x_{<t},v]$ denotes the concatenation of $v$ to the previous context. Consistent with ARGS, $w$ is set to 1.5 in our experiments.

Two decoding strategies are used in ARGS: greedy and top-k decoding. For greedy-based decoding, the token with the maximum score is chosen. For top-k decoding, tokens are sampled from a renormalized probability distribution among the top-k candidate tokens, with the following probability:
\begin{equation}
    p(v,x_{<t},\tau) = \frac{exp(s(v,x_{<t}))/\tau}{\sum_{v_i\in V}exp(s(v_i,x_{<t}))/\tau}
\end{equation}where $\tau$ is the temperature control. 
\subsubsection{GPT-4 Evaluation}
\label{appen_hh_gpt4}
Previous research \cite{zheng2024judging} has demonstrated that using GPT-4 as a proxy aligns with human evaluations over 80\% of the time for quality assessments, offering a scalable method to approximate human preferences. Following the ARGS methodology, we prompt GPT-4 to assign scores to 300 randomly sampled responses based on helpfulness, harmlessness, relevance, accuracy, and insightfulness. We show the system prompt we used for GPT-4 evaluation on the win-tie rate of ARGS and Otter against the baseline models. \begin{quote}
    \textbf{System prompt:} \emph{You are a helpful, harmless, and precise assistant for checking the quality of the answer. We would like to request your feedback on the performance of two AI assistants in response to the user question displayed above. Please rate the harmlessness, helpfulness, and level of detail of their responses. Your evaluation should consider factors such as the helpfulness, harmlessness, relevance, accuracy, depth, creativity, and level of detail of the response. Each assistant receives an overall score on a scale of 1 to 10, where a higher score indicates better overall performance. Please first output a single line containing only two values indicating the scores for Assistant 1 and 2, respectively. The two scores are separated by a space. In the subsequent line, please provide a comprehensive explanation of your evaluation, avoiding any potential bias and ensuring that the order in which the responses were presented does not affect your judgment.}
\end{quote}
\subsubsection{ARGS Evaluation Metrics}
\label{appen_hh_eval}
Diversity: This metric aggregates n-gram repetition rates across the generated texts. A higher diversity score indicates the model's capacity to produce a broad spectrum of vocabulary, enhancing the richness and variability of the outputs.

Coherence: This metric is estimated by calculating the cosine similarity between the sentence embeddings of the prompt and its continuation. We utilize the pre-trained SimCSE sentence embedding model~\cite{su2022contrastive} to obtain the embeddings, quantifying the semantic coherence between the input prompt and the generated text. 

\subsection{Toxicity Reduction}
\label{appen_toxi}
\subsubsection{Hyperparameter Setting}
The hyperparameter settings for the Otter and training process are summarized in Table~\ref{tab:appen_bi_hyper}. During the Otter training phase, we first introduce the positive expert Otter parameters and fine-tune these parameters. Subsequently, we incorporate the negative expert Otter on top of the positive expert Otter and fine-tune only the negative Otter parameters. For the generation hyperparameter settings, please refer to the following section.
\begin{table}[]
\centering
\begin{tabular}{lc}
\toprule
Parameters            & Value \\ \midrule
\# of training epochs & 3     \\
Learning rate         & 1e-4  \\
Warm-up               & 0.01  \\
LayerNorm regularization $\lambda$ & 10    \\
FFN Otter input dim   & 256   \\
FFN Otter inner dim   & 512   \\
\# Otter attn heads    & 5     \\ \bottomrule
\end{tabular}
\caption{Summary of hyperparameter settings of Otter on controlled text generation for reducing toxicity}
\label{tab:appen_bi_hyper}
\end{table}
\subsubsection{Detailed Metrics}
Toxicity metrics include the averaged maximum toxicity (avg. max.) score across 25 generations and the empirical probability of generating toxic text at least once in 25 generations, as evaluated by the Perspective API.
\subsubsection{DEXP Decoding}
The DEXP approach introduces two additional models, with the same size as the original model, to calibrate the model-generated response. Specifically, these two additional models are trained on two opposite datasets, non-toxic and toxic, respectively. Consequently, these two models are categorized as the positive expert and the negative expert. During inference, the original model's output can be calibrated by following the equation:
\begin{equation}
    p(x_t|x_{<t}) = softmax(z_t + \alpha(z^+_t-z^-_t))
\end{equation} where $z^+_t$ and $z^-_t$ are the output token probabilities of the positive and negative experts, respectively. $z_t$ is the output of the original model, and $\alpha$ is a hyperparameter set to 2.0, following the DEXP paper.

For the negative expert-only (or anti-expert-only) setting, following the DEXP approach, we simply sample tokens using the following equation:
\begin{equation}
    p(x_t|x_{<t}) = softmax((1+\alpha)z_t - \alpha z^-_t))
\end{equation}
\subsection{Speculative Decoding}
\label{appen_sd}
\subsubsection{Hyperparameter Setting}
The hyperparameter for Otter and training is shown in Table.~\ref{tab:appen_sd_hyper}. We use the same training hyper-parameters to fine-tune the TinyLlama to obtain Vicuna-draft. 
\begin{table}[]
\centering
\begin{tabular}{lc}
\toprule
Parameters            & Value \\ \midrule
\# of training epochs & 5     \\
Learning rate         & 5e-4  \\
Warm-up               & 0.01  \\
LayerNorm regularization $\lambda$ & 50    \\
FFN Otter input dim   & 128   \\
FFN Otter inner dim   & 256   \\
\# Otter attn heads    & 4     \\ \bottomrule
\end{tabular}
\caption{Summary of hyperparameter settings of Otter on speculative decoding}
\label{tab:appen_sd_hyper}
\end{table} 
\subsubsection{Speculative Decoding}
The Medusa approach trains additional language model heads (i.e. Medusa heads) that are aligned with the original language model head. This allows the newly trained Medusa heads to predict the next $k$ tokens simultaneously. Consequently, the training objective of speculative decoding is to maximize the probability of the next $i+1$ token for the $i$-th head. This can be observed from the loss function:
\begin{equation}
    \mathcal{L} = \sum_{k=1}^{K}-\lambda_k logp_t^{(k)}(y_{t+k+1})
\end{equation} where $k$ denotes the $k$-th Medusa head, and $\lambda_k$ is a hyperparameter set as the $k$-th power of a constant, typically 0.8, following the Medusa setting.

It is worth noting that in our experiment, we only adopt Medusa-1 as the baseline. This is because Medusa-2 alters the overall model output distribution, which may potentially lead to unintended consequences such as catastrophic forgetting and hallucination~\cite{jian2022embedding,pmlr-v234-zhai24a,lin2023speciality}. 

\subsubsection{Medusa vs. Draft Model}
In original speculative decoding, one or more small draft models are deployed. However, Otter followed the Medusa model approach, where we added extra decoding heads on top of the original model's last layer. The extra heads predict the next tokens in parallel, making the process more efficient than the vanilla speculative decoding method, which predicts draft tokens sequentially.

Here's how it works more specifically:
\begin{itemize}
    \item[Step 1] Initial Stage: Suppose the input token list is $\{x_1,x_2,\cdots, x_t\}$, we start with the original model's last hidden states ($H_t$) at position t.
    \item[Step 2] Draft Stage: We use the Otter-inserted $ H_t'$ to map to K new decoding heads. Each decoding head $k$ predicts the next $k+1$ tokens in parallel. Therefore, we obtained draft tokens: $\{x_t,x_{t+1},\cdots, x_{t+k+1}\}$
    \item[Step 3] Verification Stage: The new token list with the draft tokens: $\{x_1,x_2,\cdots, x_t,x_{t+1},\cdots, x_{t+k+1}\}$ is then verified by the original LLM. Suppose the first $m$ tokens are accepted, then we have the final current generation token list: $\{x_1,x_2,\cdots, x_t,x_{t+1},\cdots, x_{t+m}\}$. This new list becomes the input in step 1 for the next iteration. Note that $H_{t+m}'$ is already computed in Otter with the verification process.
\end{itemize}

In the vanilla speculative decoding setting, the inference process takes several draft model forward passes to draft and one base model forward pass for verification. However, in the Otter/Medusa setting, the inference process will only use the $k$ extra heads to draft and then take one base model forward pass for verification. Draft tokens are predicted in parallel, and the draft heads have significantly fewer parameters than draft models. This makes Otter/Medusa faster and less computationally intensive.
\section{Task Heads in Otter}
The final layer of the Otter-modified transformer outputs hidden state $\tilde{H}=[H_o:H']$, where $H_o$ is the original hidden state. For classification-related tasks, such as the helpful and harmless alignment and detoxification experiments, we have the projection layer defined as follows:
\begin{equation}
\begin{aligned}
     o_{lm} &= lm\_head(H_o)\\
     o_{otter}&=\sigma (W_{o_{n}}H')
\end{aligned} 
\end{equation}where $W_{o_{n}} \in \mathbb{R}^{1\times h'}$ is trainable matrix that projects $H'$ to a singular tensor. The original output $o_{lm}$ is obtained via the original $lm\_head$ project layer. 

For generation-related tasks, such as speculative decoding tasks, we define the projection layer as follows: 
\begin{equation}
\begin{aligned}
    o_{lm} &= lm\_head(H_o)\\
    H_{m} &= W_{h_{m}}H' \\
    o_{otter}&= lm\_head(H_{m} + H_o)
\end{aligned}
\end{equation} where $W_{h_{m}} \in \mathbb{R}^{h_o\times h'}$ is a trainable matrix to map $H'$ to $H_{m}$ that has the same hidden dimension as $H_o$.

\section{Code Modification}
\label{appen_code_modify}
We demonstrate that the entire modeling architecture of LLMs remains unchanged, with the exception of a single-line modification for RMSNorm. The following code provides the necessary adjustment to RMSNorm in the PyTorch environment.

\definecolor{dkgreen}{rgb}{0,0.6,0}
\definecolor{gray}{rgb}{0.5,0.5,0.5}

\lstset{
    language=Python,
    basicstyle=\ttfamily,
    basicstyle=\small,
    numbers=left,
    numberstyle=\tiny,
    frame=single,
    breaklines=true,
    showstringspaces=false,
    numberstyle=\tiny\color{gray},
  keywordstyle=\color{blue},
  commentstyle=\color{dkgreen}
}
\begin{lstlisting}
# Modified RMSNorm function.
def RMSNorm(hidden_states):
    # change from:
    # variance = hidden_states.pow(2).mean(-1, keepdim=True)
    # to: 
    variance = hidden_states[...,       :ori_hdim].pow(2).mean(-1, keepdim=True)

    # below lines are not changed
    hidden_states = hidden_states * torch.rsqrt(variance + variance_epsilon)
    return weight * hidden_states
\end{lstlisting}

\section{Training Time Analysis}
In general, the training for Otter is faster than the baselines, as only a few parameters need to be trained. To illustrate this difference, we provide training time for the helpful and harmless human preference alignment task. All models were trained on the same 4 NVIDIA A100 80G GPUs with DeepSpeed zero-stage 2. The per-device batch size is 1 and the gradient accumulation step is 8.
\begin{table}[]
\centering
\begin{tabular}{lcc}
\toprule
Methods & Num Params & Training Time \\ \midrule
ARGS    & 6.74B      & 163mins                 \\
Otter   & 8.51B      & 108mins                 \\ \bottomrule
\end{tabular}
\caption{The training time of ARGS and Otter comparison on preference alignment task}
\label{tab:appen_sd_hyper}
\end{table}

\section{Additional Preference Alignment Results}
Besides Llama-7b-chat already experimented in Table~\ref{tab:args_main}, here we supplemented the ARGS and Otter methods on the Llama2-7b-instruct model to evaluate their performance on the human preference alignment task.
\begin{table*}[h]
\small
\centering
\begin{tabular}{lcccccc}
\toprule
\textbf{Method}            & \textbf{Average Reward $\uparrow$} & \textbf{Diversity ↑} & \textbf{Coherence ↑} & \textbf{Time Overhead ↓} & \textbf{Space Overhead ↓} \\ \midrule
Greedy                     & 4.468                     & 0.583                & 0.454                & 1.0x                     & 1.0x                     \\
Greedy+ARGS                & \textbf{5.221}            & 0.617                & 0.461                & 2.06x                    & 2.04x                    \\
Greedy+Task Head Only      & 4.536                     & 0.542                & 0.497                & 1.01x                    & 1.03x                    \\
Greedy+Otter               & 5.175                     & 0.739                & 0.515                & \textbf{1.03x}           & \textbf{1.25x}           \\ \midrule
Top-k                      & 4.112                     & 0.695                & 0.479                & 1.0x                     & 1.0x                     \\
Top-k+ARGS                 & 4.819                     & 0.702                & 0.486                & 2.16x                    & 2.04x                    \\
Top-k+Task Head Only       & 4.360                     & 0.689                & 0.472                & 1.02x                    & 1.03x                    \\
Top-k+Otter                & 4.807                     & 0.820                & \textbf{0.528}       & 1.03x                    & 1.25x                    \\ \midrule
Top-p                      & 4.146                     & 0.624                & 0.417                & 1.0x                     & 1.0x                     \\
Top-p+ARGS                 & 4.876                     & 0.669                & 0.445                & 2.18x                    & 2.04x                    \\
Top-p+Task Head Only       & 4.297                     & 0.628                & 0.458                & 1.02x                    & 1.03x                    \\
Top-p+Otter                & 4.864                     & \textbf{0.783}       & 0.507                & 1.04x                    & 1.25x                    \\ \bottomrule
\end{tabular}
\caption{Comparison of ARGS and Otter using Llama2-7b-chat}
\label{tab:addi_comp}
\end{table*}
As shown in Table~\ref{tab:addi_comp}, the Llama2 model itself performs better than Llama in terms of generating more helpful and harmless responses (Llama-7b obtained 3.981 and 3.757 average rewards), both ARGS and Otter can further enhance the model's performance, leading to higher average reward. More importantly, Otter significantly reduced time, and space overhead while keeping the same performance as ARGS.

\end{document}